\documentclass[journal]{IEEEtran}
\usepackage{bm}

\ifCLASSINFOpdf
\usepackage[pdftex]{graphicx}
\else
\usepackage[dvips]{graphicx}
\def\IEEEkeywordsname{Keywords}
\DeclareRobustCommand*{\IEEEauthorrefmark}[1]{\raisebox{0pt}[0pt][0pt]{\textsuperscript{\footnotesize\ensuremath{\ifcase#1\or *\or \mathsection\or %
    \mathparagraph\or 1\or 2\or 3\or 4\or \|\or **\or \dagger\or \ddagger\or \dagger\dagger %
    \or \ddagger\ddagger \else\textsuperscript{\expandafter\romannumeral#1}\fi}}}}
\fi

\begin{document}
\bstctlcite{IEEEexample:BSTcontrol}
\title{Deformation estimation of an elastic object\\ by partial observation using a neural network}
%\author{Utako~Yamamoto\IEEEauthorrefmark[1],~%\IEEEmembership{Member,~IEEE,}
%        Megumi~Nakao,~%\IEEEmembership{Member,~IEEE,}
%        Masayuki~Ohzeki,~%\IEEEmembership{Member,~IEEE,}
%        and~Tetsuya~Matsuda%,~\IEEEmembership{Member,~IEEE}% <-this % stops a space
%        
%\thanks{This research was supported by 
%the Center of Innovation (COI) Program from the Japan Science and Technology Agency (JST) 
%and Japan Society for the Promotion of Science (JSPS) Grant-in-Aid for Young Scientists (B) Grant Number JP16K16407.}
%\thanks{U. Yamamoto, M. Nakao and T. Matsuda are with 
%the Department of Systems Science, Graduate School of Informatics, Kyoto University, 
%Yoshida-honmachi, Sakyo-ku, Kyoto-city, Kyoto, 606--8501, Japan. 
%(e-mail: utako@i.kyoto-u.ac.jp).}% <-this % stops a space
%\thanks{M. Ohzeki is with 
%the Department of Applied Information Sciences, Graduate School of Information Sciences, Tohoku University, 
%6-3-09 Aoba, Aramaki aza, Aoba-ku, Sendai-city, Miyagi, 980--8579, Japan.}}% <-this % stops a space

\author{\IEEEauthorblockN{Utako Yamamoto\IEEEauthorrefmark{2}\IEEEauthorrefmark{1}, 
        Megumi Nakao\IEEEauthorrefmark{2}\IEEEauthorrefmark{4}, 
        Masayuki Ohzeki\IEEEauthorrefmark{3}\IEEEauthorrefmark{5}, and 
        Tetsuya Matsuda\IEEEauthorrefmark{2}\IEEEauthorrefmark{6}}
        \\[10pt]

\IEEEauthorblockA{\IEEEauthorrefmark{2}Department of Systems Science, Graduate School of Informatics, Kyoto University, \\
Yoshida-honmachi, Sakyo-ku, Kyoto-city, Kyoto, 606--8501, Japan} 

\IEEEauthorblockA{\IEEEauthorrefmark{3}Department of Applied Information Sciences, Graduate School of Information Sciences, Tohoku University, \\
6-3-09 Aoba, Aramaki aza, Aoba-ku, Sendai-city, Miyagi, 980--8579, Japan}

\IEEEauthorblockA{\IEEEauthorrefmark{1}Corresponding author. E-mail: utako@i.kyoto-u.ac.jp, Tel.: +81-75-753-4940, fax: +81-75-753-3375.}

\IEEEauthorblockA{\IEEEauthorrefmark{4}E-mail: megumi@i.kyoto-u.ac.jp}
\IEEEauthorblockA{\IEEEauthorrefmark{5}E-mail: mohzeki@tohoku.ac.jp}
\IEEEauthorblockA{\IEEEauthorrefmark{6}E-mail: tetsu@i.kyoto-u.ac.jp}

%\thanks{Corresponding author: U. Yamamoto (E-mail: utako@i.kyoto-u.ac.jp, Tel.: +81-75-753-4940, fax: +81-75-753-3375.)}
}

\markboth{}%
{Utako \MakeLowercase{\textit{et al.}}: Deformation estimation of an elastic object by partial observation using a neural network}

\maketitle

\begin{abstract}
Deformation estimation of elastic object assuming an internal organ is important for the computer navigation of surgery. 
The aim of this study is to estimate the deformation of an entire three-dimensional elastic object 
using displacement information of very few observation points. 
A learning approach with a neural network was introduced to estimate the entire deformation of an object. 
We applied our method to two elastic objects; a rectangular parallelepiped model, 
and a human liver model reconstructed from computed tomography data. 
The average estimation error for the human liver model was 0.041 mm 
when the object was deformed up to 66.4 mm, from only around 3 \% observations. 
These results indicate that the deformation of an entire elastic object can be estimated with an acceptable level of error 
from limited observations by applying a trained neural network to a new deformation. 
\end{abstract}

\begin{IEEEkeywords}
Deformation estimation, Multi-layer neural network, Partial observation, Nonlinear finite element method, Elastic object, Liver.
\end{IEEEkeywords}

\IEEEpeerreviewmaketitle

%\newpage

%||||||||||||||||||||||||||||||||||||||||||||||||||||||||||||||||||||||||||||||||||||||
\section{Introduction}
\IEEEPARstart{T}{he} 
number of diseases that can be treated by surgical operation has increased in recent years 
because of the development of medical technology. 
However, the complexity of treatment plans means that surgical navigation 
under treatment is a potentially valuable development. 
In surgical navigation, computer simulation of treatment plans is constructed before a surgery, 
and a virtual internal organ deformed following surgical treatment is presented to the surgeon during the surgery. 
It helps surgeons to operate as planned and 
provides an alert to prevent accidents, for example, they hurt important blood vessels.
In surgical operations, laparoscopic surgery is considered less trauma for patients than laparotomy, 
and leads to faster recovery \cite{Shu2013}. 
However, laparoscopic surgery is more difficult for doctors than laparotomy surgery, 
because it involves restricted vision and limited freedom of surgical instruments, 
making it difficult to move the instruments in the operating field \cite{Fuentes2014}. 
Consequently, the surgical navigation is particularly valuable in laparoscopic surgery. 
It is said that surgeons feel comfortable with the procedures after they have completed procedures on 12-18 patients \cite{Meadows2002}. 
Surgical navigation is thought to reduce the risk of mistakes before the doctor gets used to treatment.
Notwithstanding the above-mentioned benefits, 
it is necessary to overcome some limits to apply the navigation to actual surgery. 
The major limitation is that it is difficult to detect the deformation of an organ 
as it is treated by a doctor and to similarly deform the organs in the simulation. 

Two are the problems that must be addressed when 
detecting and implementing the deformation of internal organs to perform the navigation during surgery. 
The first is to track the deformation of the actual organ surface. 
Many surface of the organ is textureless, 
and it is difficult to specify the position of the surface point moved before and after the deformation \cite{Malti2014}. 
The second task is to obtain deformation information of the entire organ from partial observation information. 
There are few situations where whole organ is visible during surgery, 
and due to the narrow field of view and the short distance to the laparoscopic camera, 
there are very few observable parts \cite{Qin2013}. 
Even if we can track the feature points on the surface, 
only the information on a part of the surface is obtained, 
and deformation information on the back side and inside can not be obtained. 
As a method to solve the first problem, 
a method combining sparse feature-matching algorithm with probabilistic modelling of Gaussian Process Regression \cite{Puerto2014} 
and a method using blood vessel features on tissue surfaces as a set of image features \cite{Lin2015} have been proposed. 
Although it is not possible to track the deformation of every position of the object, 
some feature point on the surface can be tracked with a certain degree of reliability.
Therefore, we believe that the second is more challenging problem. 

We approach the problem of obtaining the entire deformation from partial observation information 
as introducing machine learning techniques; especially neural networks \cite{Hagan2014}. 
In many cases, it is difficult to install equipment that measures the force pulling an organ during surgery, 
and only information on the movement of the surface point caused by deformation is obtained. 
We argue that, estimating the overall deformation from observation information of partial displacement 
is an important step in presenting the state of an organ in surgical navigation in accordance with the treatment of a doctor. 
It contributes to providing the deformation information of the part that the doctor can not observe directly, 
even if the visible area is narrow and there are few points that can be tracked. 
The focus of this research is hence to estimate the entire deformation from deformation information of very few surface observation 
when the force applied to the elastic body is unknown. 

\begin{figure}[!t]
\centering
\includegraphics[width=80mm]{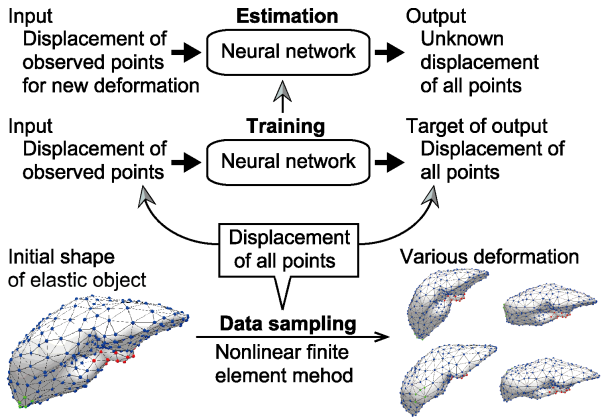}
\caption{Framework of this study. 
After data sampling of displacement with nonlinear finite element method 
for elastic object assuming internal organ, 
the neural network is trained with the data set to estimate the displacement of all points 
including unobservable points from the displacement information of very few observable points.}
\label{fig:Framework}
\end{figure}

Since neural networks have many parameters to be optimized, 
they have high expression power \cite{Kecman2001}. 
The neural network can be trained via back-propagation method \cite{LeCun1989}. 
It was reported that neural network for document recognition had an outstanding performance on large amount of training data \cite{Lecun1998}. 
However, in the field of medical engineering, these advances have been used in few cases. 
In the current study, we propose a method for estimating the displacement of an entire elastic object 
from the displacement measured at very few observable points by training a neural network to learn from deformation data 
calculated by the simulation under the condition that the initial shape is known. 
No study we are aware of has addressed the use of neural network in deformation estimation from only displacement information 
and our study will be the first. 
The framework of this study is shown in Fig. \ref{fig:Framework}.
In this paper, we investigated the size of the neural network and iteration number of optimization 
suitable for deformation learning. 
We also investigated the generalization capability of our method by use of deformation data set with multiple pattern. 
Our results indicate that our proposed method can estimate the whole deformation within the tolerance range assumed for surgery support. 

In this paper, we first describe the method for calculating the three-dimensional shape 
of the deformed elastic object using the nonlinear finite element method, in Section \ref{sec:FEM}, 
after we mention the related works in Section \ref{Related}. 
We then describe the algorithm of the deformation estimation method using the neural network in Section \ref{sec:NN}. 
In Section \ref{sec:ObtainData}, we describe the method for acquiring the experimental data 
using the nonlinear finite element method of the elastic object, 
and Section \ref{sec:Result} reports the results of applying the deformation estimation 
with the neural network to experimental data. 
Finally, we discuss the implications of these findings in Section \ref{sec:Discussion}, 
then summarize the study in Section \ref{Fin}.

%||||||||||||||||||||||||||||||||||||||||||||||||||||||||||||||||||||||||||||||||||||||
\section{Related work}
\label{Related}
The finite element method is commonly used as a technique for calculating the deformation 
of a living body, based on a model \cite{Belytschko2014}. 
Many attempts have been made to estimate the deformation of a living body 
by applying the finite element method to the visualization of organs, 
training of doctors and surgical planning. 
Using computed tomography (CT) \cite{FCAMRT2015} before surgery, 
attempts to simulate and visualize operations such as grasping or incising organs 
during surgical operations have been reported \cite{Nakao2010}. 
In addition, simulators for doctors to practice surgical suturing procedures have been developed \cite{Berkley2004}. 
One study reported a model for deforming organs in accord 
with respiration for dynamic radiation irradiation planning, constructed using the finite element method \cite{Nakao2007}.

As a method for improving the accuracy of deformation estimation 
using the finite element method to increase calculation speed, 
a technique was reported that the rotational elements of an object were extracted 
from deformation, and it computed only with elements other than rotation \cite{Muller2002}. 
In addition, one study reported a method for simplifying an object shape by expressing the element 
with a pair of edges that shared an element between the elements that divided the object \cite{Kikuuwe2009}. 
In addition, techniques for shape matching based on a dynamic deformation model 
have been proposed for surgical assistance \cite{Suwelack2014}. 
However, in each of these cases, pre-existing knowledge of the mechanical properties of the object 
(e.g., hardness) is necessary for estimating deformation.

In contrast, deformation was simulated in one study while applying force to an organ using a neural network \cite{Morooka2010,Morooka2012,Morooka2013}, 
as a method for estimating deformation based on accumulated deformation data and machine learning. 
In a study by Morooka et al., 
deformation data for a deformed object were obtained beforehand by applying many types of force using the finite element method, so that the
relation between the external force applied to the organ 
and the compressed vertex coordinates of the mesh could be learned by a neural network. 
This method has the advantage of not requiring knowledge of the mechanical properties 
of organs before the time of deformation estimation. 
However, in this method it is still necessary to input the vertex coordinates of the contact points 
and the direction and magnitude of the force.

A recent report described the development of a method 
for reconstructing the three-dimensional shape of an organ 
from an endoscopic image, to enable navigation at the time of laparoscopic surgery \cite{MaierHein2013}. 
However, in this approach, it is difficult to extract feature points corresponding to 
before and after deformation from the limited range of observable areas with high precision. 
In order to estimate the shape of the elastic object, 
matching algorithms based on shape descriptors \cite{Santos2014}, 
a method using weighted patch iterative closest point of the mesh \cite{Simpson2012}, 
and a method using laser range scanner data to obtain information about the surface of the moving tissue \cite{Dumpuri2010} have been proposed. 
In another study, the external force applied to an object was estimated 
using partial observation, to overcome limited visibility 
at the time of surgical treatment \cite{Sakata2017}. 
In their work, it is necessary to assume that the properties related to the shape of the object are invariant 
before and after the deformation. 
In addition, attempts to estimate organ deformation from several silhouette images obtained 
from objects have been reported \cite{Saito2015}. 
However, there are limitations on the accuracy of estimation using this method, 
because there is little information for determining deformation.

Although previous studies have attempted to use neural networks to estimate the force applied to an elastic body 
from images of the deformed elastic body \cite{Greminger2003,Aviles2014}, 
these methods require observation of the whole object. 
In the field of cell Micromanipulation, studies also have been reported to estimate the force using a neural network \cite{Karimirad2013,Karimirad2014}. 
Moreover, a study examining the estimation of the outline of an object using a neural network 
demonstrated the advantage of not requiring a priori knowledge on the material of an object \cite{Cretu2008,Cretu2010}. 
However, this method requires information about the magnitude of the force deforming the object 
and the position of the contact point, 
making it unsuitable for surgical operations 
where measuring the magnitude of the force applied by an instrument to an organ is difficult.

%||||||||||||||||||||||||||||||||||||||||||||||||||||||||||||||||||||||||||||||||||||||
\section{Elastic deformation with the nonlinear finite element method}
\label{sec:FEM}
%In the finite element method, deformation of the structure can be calculated 
%by representing the relationship between force and displacement 
%when a force is applied to a truss structure with which elements are connected via a node 
%by an equation using a stiffness matrix. 
%By dividing the continuous elastic body into a triangle in two dimensions 
%or a tetrahedron element in three dimensions and treating it as a truss structure 
%with a connection point between the elements, 
%deformation can be represented by the equation with a matrix. 

We employed the nonlinear finite element method with forced displacement boundary conditions 
to obtain data samples in the current study. 
When the boundary conditions to deform the elastic object are known, 
deformation of the structure can be calculated by the nonlinear finite element method, 
which has been developed over the years. 
We express the calculation method in this section. 

%----------------------------------------
\subsection{Nonlinear finite element method}
The relationship between force and displacement can be expressed by the following stiffness equation, 
assuming that the three-dimensional elastic body is divided into tetrahedral meshes. 
The nonlinear deformation is expressed by this linear equation during micro deformation 
and calculated nonlinearly by repeating the micro deformation. 
Here, the force applied to the vertex of the mesh is $\bm{f}$, 
and the displacement for each vertex is $\bm{u}$. 

\begin{equation}
\bm{f} = K \bm{u}
\label{eq:fKu}
\end{equation}

The stiffness matrix $K$ is a matrix determined 
by the shape and mechanical properties of the elastic body. 
In the nonlinear finite element method, 
$K$ is updated according to the shape every time the elastic body is slightly deformed. 
The stiffness matrix $K$ of the entire mesh can be formed 
by overlapping the element stiffness matrix $K_{e}$ of each mesh corresponding to the vertex 
shared by different meshes. 
From the matrix $B$ representing the shape of one mesh, 
the matrix $D$ representing the mechanical properties 
and the volume $\Delta$ of the mesh, $K_{e}$ can be defined as follows.

\begin{equation}
K_{e} = \Delta B^{T} D B
\label{eq:BTB}
\end{equation}

Assuming that the deformation does not change the mechanical properties of the elastic body, 
$K_{e}$ is updated by changing $\Delta$ and $B$ caused by deformation of the mesh. 
For a mesh structure whose initial shape is already known, 
the initial $K$ is obtained from the formula (\ref{eq:BTB}), 
and the micro deformation generated when a minute force or minute displacement 
is applied to the mesh is calculated using the equation (\ref{eq:fKu}), 
and $K$ after deformation is obtained by the formula (\ref{eq:BTB}). 
By repeating this calculation, 
we can obtain the displacement of all vertices of the mesh structure  
and the force applied to the contact points after a large deformation.

%----------------------------------------
\subsection{Deformation by forced displacement boundary conditions}
To deform the elastic object, 
a constraint condition is required for setting $N_{f}$ vertices among all vertices $N_{a}$ 
of the mesh to a fixed point that does not move. 
This is because, even if a force is applied to a structure without a fixed point, 
the whole shape does not change and only translates or rotates. 
Even if the row or column corresponding to the fixed point part is excluded 
in the expression (\ref{eq:fKu}), 
the expression (\ref{eq:fKu}) holds depending on the characteristics of the matrix. 
Among the $N$ vertices excluding the fixed point, 
$N_{c}$ vertices are selected as contact points to be manipulated for deforming the whole object.

When using forced displacement as a boundary condition of deformation, 
we first rewrite the equation (\ref{eq:fKu}) with $L$ as the inverse matrix of $K$ 
to calculate the finite element method by giving displacement $\bm{u}$.
\begin{equation}
\bm{u} = K^{-1} \bm{f} = L \bm{f}
\label{eq:uLf}
\end{equation}

The displacement $\bm{u}_{c}$ of the contact point is known 
because it is given as a boundary condition, 
but the displacement $\bm{u}_{n}$ of the vertices other than the contact point 
among the vertices excluding the fixed point is unknown. 
By dividing the vector $\bm{u}$ into known and unknown elements, 
equation (\ref{eq:uLf}) can be rewritten as follows: 
$L$ is known since we use the value of $L$ before micro deformation. 
The subscript $c$ and $n$ signify the vertices of contact points and the vertices except for contact and fixed points, respectively. 

\begin{equation}
\left(
\begin{array}{c}
 \bm{u}_{c} \\
 \bm{u}_{n}
\end{array}
\right)
= 
\left(
\begin{array}{cc}
L_{cc} & L_{cn} \\
L_{nc} & L_{nn}
\end{array}
\right)
\left(
\begin{array}{c}
 \bm{f}_{c} \\
 \bm{f}_{n}
\end{array}
\right)
\label{eq:uLfcn}
\end{equation}

First, using only the part related to $\bm{u}_{c}$ that is known, 
the force $\bm{f}$ applied to all vertices is obtained as follows:

\begin{equation}
 \bm{u}_{c} = 
\left(
\begin{array}{cc}
L_{cc} & L_{cn}
\end{array}
\right)
\left(
\begin{array}{c}
 \bm{f}_{c} \\
 \bm{f}_{n}
\end{array}
\right)
\label{eq:uLfc}
\end{equation}

Second, the displacement $\bm{u}_{n}$ of unknown vertices is calculated using the obtained force $\bm{f}$ as follows: 

\begin{equation}
 \bm{u}_{n} = 
\left(
\begin{array}{cc}
L_{nc} & L_{nn}
\end{array}
\right)
\left(
\begin{array}{c}
 \bm{f}_{c} \\
 \bm{f}_{n}
\end{array}
\right)
\label{eq:uLfn}
\end{equation}

With the equations described above, 
it is possible to obtain the displacement of all vertices
other than the contact point, with the forced displacement boundary condition given to the contact point.

%||||||||||||||||||||||||||||||||||||||||||||||||||||||||||||||||||||||||||||||||||||||
\section{Methods}
\label{sec:NN}
In the current study, a neural network was used for learning and estimation of elastic deformation. 
In this section, we describe the structure, cost function, optimization method, 
and input/output data of the neural network used.

%----------------------------------------
\subsection{Neural network}
\begin{figure}[!t]
\centering
\includegraphics[width=85mm]{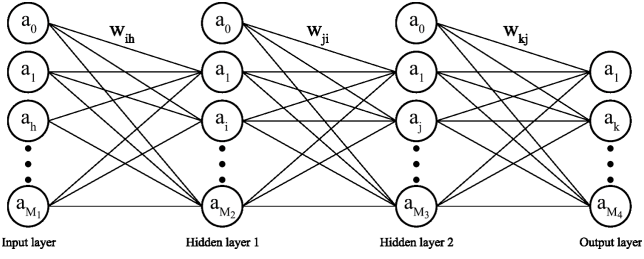}
\caption{Neural network with two hidden layers.
In each layer, the circle represents the node and $a$ represents the value of the node, 
and the line connecting the nodes represents the connection weight $w$ between the layers.}
\label{fig:NN}
\end{figure}

In this study, a neural network with two hidden layers was adopted, as shown in Fig. \ref{fig:NN}. 
Let $M_{1}$，$M_{2}$，$M_{3}$，$M_{4}$ be the number of nodes in each layer from the input side, 
and let the values of nodes in each layer be $a_{h}$，$a_{i}$，$a_{j}$，$a_{k}$. 
$a_{0}$ is a bias term, and the value is 1. 
Let $w_{ih}$，$w_{ji}$，$w_{kj}$ be the weights for converting the feature quantity 
when proceeding to the next layer. 
The rectified linear unit (ReLU) \cite{Nair2010} was used as the activation function in the middle layer. 
\begin{equation}
g(z) = \left\{ 
\begin{array}{cc}
0 & (z < 0) \\
z & (z \geq 0)
\end{array}
\right.
\label{eq:ReLU}
\end{equation}

With $a_{h}$ as input, the value of the node at each layer can be expressed as follows: 
\begin{eqnarray}
a_{i} &=& g(z_{i}),\:  z_{i} = \sum_{h=0}^{M_{1}} w_{ih} a_{h} \\
a_{j} &=& g(z_{j}),\:  z_{j} = \sum_{i=0}^{M_{2}} w_{ji} a_{i} \\
a_{k} &=& \sum_{j=0}^{M_{3}} w_{kj} a_{j}
\label{eq:NN}
\end{eqnarray}

The number of nodes in the input and output layers is determined 
by the dimensions of the input/output data, 
but the number of nodes of the two hidden layers can be arbitrarily selected. 
Since the input/output data is a real number with a negative value, 
we did not set the activation function in the output layer, but only multiplied 
and added the feature quantity in the second hidden layer.

%----------------------------------------
\subsection{Cost function for Learning}
The error function between the output $\bm{a}$ of the neural network 
and the target data $\bm{y}$ which means correct solution data is defined by the L2 norm.
\begin{equation}
E(\bm{a}) = \frac{1}{2} || \bm{a} - \bm{y} ||_{2}^{2}
\label{eq:error}
\end{equation}

To prevent over-fitting, 
all three weights were subjected to regularization by the L2 norm. 
Therefore, the cost function at the output layer is expressed as follows. 
Here, $m$ is the number of training data, $d$ is the data number, 
and $n_{1},n_{2},n_{3}$ represent the number of elements of the weight matrices $w_{ih},w_{ji},w_{kj}$ 
excluding the part related to the bias term.
\begin{eqnarray}
J = &&\frac{1}{m} \sum_{d=1}^{m} \sum_{k=1}^{M_{4}} 
\left\{ \frac{1}{2} ( a_{k}^{(d)} - y_{k}^{(d)} )^{2}\right\} 
  + \frac{\lambda_1}{2n_{1}} \sum_{h=0}^{M_{1}} \sum_{i=1}^{M_{2}}  (w_{ih})^2 \nonumber \\
&&+ \frac{\lambda_2}{2n_{2}} \sum_{i=0}^{M_{2}} \sum_{j=1}^{M_{3}}  (w_{ji})^2 
+ \frac{\lambda_3}{2n_{3}} \sum_{j=0}^{M_{3}} \sum_{k=1}^{M_{4}}  (w_{kj})^2 
\label{eq:J}
\end{eqnarray}

Regularization with the L2 norm prevents the elements of the weight 
that are optimized parameters from becoming too large. 
Since learning is performed without unnecessarily increasing the elements' value of weight, 
this prevents over-fitting that would make the optimized neural network match 
only the training data. 
$\lambda$ is a regularization coefficient with a positive value, 
and increasing $\lambda$ strengthens the effect of suppressing the weight factor 
from becoming a large value.

To optimize the weight $w$ using the error back propagation method 
and minimize the above cost function, 
the derivative of the cost function was obtained. 
Since the error function in the output layer is defined by the L2 norm, as shown in equation (\ref{eq:error}), 
and the activation function is not set in the output layer, 
the differentiation of the error in the output layer is expressed as follows:

\begin{eqnarray}
\delta_k = a_k - y_k
\label{eq:diff}
\end{eqnarray}

The differential function of ReLU is: 
\begin{equation}
g{'}(z) = \left\{ 
\begin{array}{cc}
0 & (z < 0) \\
1 & (z > 0)
\end{array}
\right.
\label{eq:ReLU_diff}
\end{equation}
where ${d g(z)}/{d z} = 0$ when $z=0$. 

Therefore, the differentiation of the cost function in each layer is as follows.
\begin{eqnarray}
\frac{\partial J}{\partial w_{ih}} &=& \frac{1}{m} \sum_{d=1}^{m} \left( \delta_k^{(d)} w_{kj} g{'}(z_{j}) w_{ji} g{'}(z_{i}) a_{h}^{(d)} \right) + \frac{\lambda_1}{n_1} w_{ih} \nonumber \\ \\
\frac{\partial J}{\partial w_{ji}} &=& \frac{1}{m} \sum_{d=1}^{m} \left( \delta_k^{(d)} w_{kj} g{'}(z_{j}) a_{i}^{(d)} \right) + \frac{\lambda_2}{n_2} w_{ji} \\
\frac{\partial J}{\partial w_{kj}} &=& \frac{1}{m} \sum_{d=1}^{m} \left( \delta_k^{(d)} a_{j}^{(d)} \right) + \frac{\lambda_3}{n_3} w_{kj} 
\label{eq:BackProp}
\end{eqnarray}

%----------------------------------------
\subsection{Optimization}
For optimization, we used a mini batch learning method, 
which is a type of stochastic gradient descent method. 
We divided all $m$ training data evenly into mini batches containing $m_1$ pieces of data 
and optimized each mini batch as a data group. 
After optimizing all mini batches, 
we shuffled all the training data and optimized it again, 
and repeated this optimization for several epochs. 
If the number $m$ of data is not equally divisible by $m_1$ pieces of data, 
the fractional data is ignored. However, because the data is shuffled for each epoch, 
the data to be ignored differs from epoch to epoch and all training data is used for optimization.

We employed Adam \cite{Kingma2014} as a stochastic gradient descent method to speed up learning. 
The Adam method does not adopt the same learning rate for all of the optimization parameters, 
but employs the learning rate that is adjusted for each parameter. 
Every time learning is iterated, the learning rate is decreased 
in the direction of the parameter with large variance, 
and the learning rate is increased in the direction of the parameter with small variance. 
This method allows the network to escape quickly from the saddle point, 
which is the main problem leading to learning stagnation. This provides the advantage of rapid convergence.

The updating equation of parameter $\theta$ in Adam is expressed as follows. 
Here, $t$ is the iteration number, 
and the estimated value obtained by bias correction of the first moment 
and the second moment bias of the slope are $\hat{m}_t$ and $\hat{v}_t$, respectively.
\begin{eqnarray}
\theta_{t} = \theta_{t-1} - \frac{\alpha \cdot \hat{m}_t}{\sqrt{\hat{v}_t} + \epsilon}
\label{eq:Adam}
\end{eqnarray}
The value of $\epsilon$ was set to $10^{-8}$, and the parameters used for calculating $\hat{m}_t$ and $\hat{v}_t$ were set 
to $\beta_1 = 0.9$ and $\beta_2 = 0.999$, according to the guidelines described by the author of the Adam method \cite{Kingma2014}. 

If the value of $\alpha$, which contributes to the overall learning rate, is kept constant, 
although the cost function decreases greatly at the beginning of optimization 
with a large value and exerts a positive effect, 
in the latter half of learning the parameter change is too large 
and the cost function does not settle on a particular value. 
Conversely, for small values of $\alpha$, the cost function decreases steadily and settles down consistently, 
but convergence occurs too slowly. 
Therefore, to substantially change the value of the parameter in the initial stage 
of learning and reduce parameter change as convergence is approached, 
$\alpha$ is set as follows so that it attenuates for each epoch:
\begin{eqnarray}
\alpha = \frac{1}{\gamma \cdot n_{e}}
\label{eq:Alpha}
\end{eqnarray}
$n_{e}$ is the epoch number, 
and the constant $\gamma$ is determined by trial and error, 50. 
The results confirmed that convergence was established 
with a smaller number of iterations compared with when $\alpha$ was constant.

%----------------------------------------
\subsection{Input and output}
In the training phase, 
$N_o$ vertices in $N$ vertices excluding fixed points were assumed to be observable points 
where displacement could be observed. 
It is not necessary for the observation point to include a point of action that applies force. 
The initial shape of the elastic object and the displacement from the initial position 
of the observation point when the elastic object was deformed were known, 
and only the displacement of the observation point was taken as the input of the neural network. 
Since the elastic object had a three-dimensional shape and underwent deformation in three-dimensional space, 
the displacement also had a component in three-dimensional space. 
Therefore, the three-dimensional axial direction component of the displacement for the observation point was input to the input layer. 
That is, the number of nodes $M_1$ of the input layer in Fig. \ref{fig:NN} is $N_o \times 3$. 
Apart from the displacement of observation point, a node was added that always input 1 as a bias term.

During training, $N$ vertex displacements including $N_u$ non-observation points were given 
as target data $\bm{y}$ for the output of the neural network. 
The number of $N$ represents the number of vertices to be estimated the displacements. 
Since the non-observation point displacement also had a three-dimensional axial direction component, 
$N \times 3$ was $M_{4}$, as shown in Fig. \ref{fig:NN}.

In the estimation phase, 
the displacement of the observation point is known and the displacement of the non-observation point 
were unknown. 
The displacement of the observation point for the new deformation is input 
in the neural network trained in the training phase, 
and we obtain the estimated value of the displacement of all vertices in the output layer. 
The position of the observation point in the whole object was required to be the same 
in the training and estimation phases. 
Since we assumed that the feature points having a characteristic shape or texture were selected as the observation points, 
the observable points could be determined in advance.

%||||||||||||||||||||||||||||||||||||||||||||||||||||||||||||||||||||||||||||||||||||||
\section{Data acquisition}
\label{sec:ObtainData}
\subsection{Model of the elastic object}

\begin{figure}[!t]
\centering
\includegraphics[width=80mm]{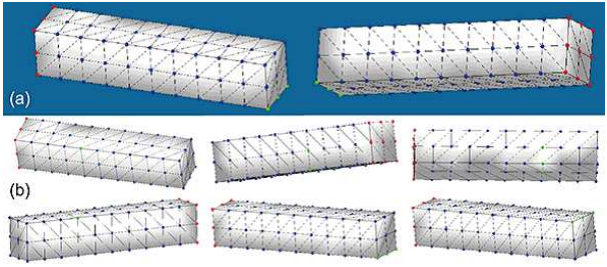}
\caption{A rectangular parallelepiped elastic object represented by a tetrahedral mesh. 
The vertices shown in red and green indicate the fixed points and the contact points, respectively. 
(a) designates a model when there was one area of contact points. 
The two figures show views from different angles. (RPP1) 
The cases in which there were six regions of contact points are shown in (b). 
Green vertices in each figure of (b) represent contact points in each case. (RPP6)}
\label{fig:EO1}
\end{figure}

In this study, 
we considered elastic bodies with rectangular parallelepiped shapes divided into tetrahedral meshes, 
as shown in Fig. \ref{fig:EO1} for testing the effectiveness of the method. 
The long side of the rectangular parallelepiped was 256 mm, 
the short sides were each 51.2 mm, 
and the interval of the closest vertex of the tetrahedral mesh was 25.6 mm. 
In the simulation space, 256 mm in real space was standardized to 1,
corresponding to the ratio for the liver model mentioned below. 
The tetrahedral mesh was evenly distributed not only on the surface but also inside the elastic object. 
There were a total of 99 vertices on the mesh, 
and the nine vertices shown in red on the edge of Fig. \ref{fig:EO1} were fixed points.

\begin{figure}[!t]
\centering
\includegraphics[width=80mm]{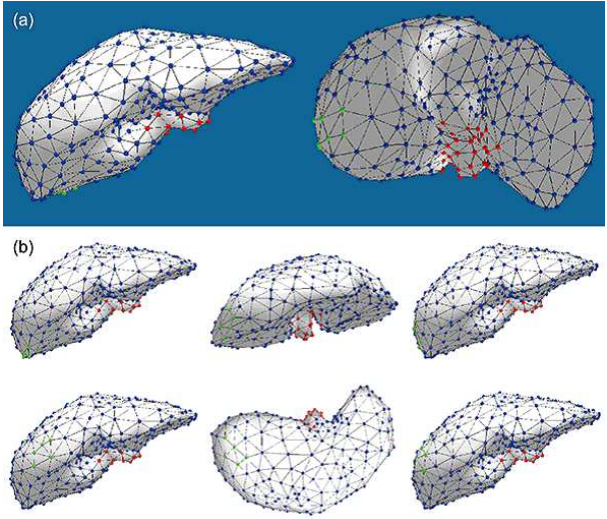}
\caption{Initial shape of liver mesh structure constructed from CT data. 
The color of the vertices was the same as for the rectangular parallelepiped shape in Fig. \ref{fig:EO1}. 
(a) A case when the region of contact points was single area. (Liver1) 
(b) Each figure represents the case of each area of contact points. (Liver6) }
\label{fig:EO2}
\end{figure}

\begin{figure}[!t]
\centering
\includegraphics[width=80mm]{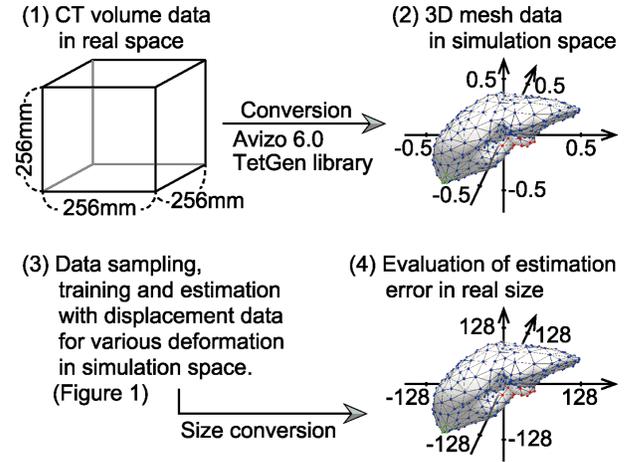}
\caption{Schematic diagram of data processing for human liver model. 
The data sampling and the training and estimation with neural network were performed in the size of simulation space, 
where the size of 256 mm in the real space was normalized to 1. 
We evaluate the estimation error in the size of real space. }
\label{fig:SchematicDiagram}
\end{figure}

In addition, to test our proposed method with an actual organ, 
the liver shown in Fig. \ref{fig:EO2} was also examined as an elastic object. 
This 3D mesh structure was constructed from human CT data. 
Since the CT data was three-dimensional volume data, 
the liver region was extracted from the whole CT volume data and converted into a surface mesh data structure 
using commercially available Avizo 6.0 software (FEI). 
A three-dimensional tetrahedral mesh structure was then constructed using the TetGen library \cite{Si2015}. 
A tetrahedral mesh was constructed so that the sizes of the triangles making up the mesh 
were as even as possible and the shape of the triangle was as close as possible to an equilateral triangle. 
The mesh contained a total of 329 vertices. 
Since the part connected to the descending aorta and the inferior vena cava does not deform greatly 
in the liver of the living body, 34 vertices (shown in red) in the mesh were used as fixed points. 
The schematic diagram presenting data processing is shown in Fig. \ref{fig:SchematicDiagram}.

The CT data were cubic voxels with 1 mm sides, 
resolution was $256\times256\times256$, and the length along the long axis of the liver was 221.3 mm. 
Since the imaging range of the CT data was normalized to 1 in the simulation space, 
256 mm in real space was 1 in the simulation space. 
Although one side of the tetrahedral mesh varied depending on the position, 
it was approximately 15 mm in real space.

%----------------------------------------
\subsection{Data sampling}
As described in Section \ref{sec:FEM}, 
we deformed the target elastic object using the nonlinear finite element method 
with a forced displacement boundary condition to obtain vertex displacement data for the experiment. 
For each target elastic object, 
a force was applied to the contact point to deform the contact point to the sample point. 
Young's modulus and Poisson's ratio, 
which represent the mechanical properties of the elastic object, 
were fixed to 1.0 MPa and 0.40, respectively, for all meshes \cite{Nakao2014}.

In obtaining deformation of an entire elastic object using the nonlinear finite element method, 
a large displacement of contact points in one calculation cycle 
can cause substantial calculation error in the stiffness equation. 
Therefore, when moving the contact point to the target sample point, 
the distance from the initial position to the sample point was divided by 1,000, and 
the force and the stiffness matrix were renewed each time a minute displacement was applied. 
Finally, the calculation error after a large deformation was ignored 
because it was negligibly small. 
It took approximately 1 minute for the rectangular parallelepiped model, 
and approximately 3 minutes for the liver model to undergo each sample of displacement after a large deformation. 

The sampling pattern of the data consisted of 
a case of sampling with a single area of contact points, 
and a case where sampling was performed for each area when there were six areas of contact points. 
In Figs. \ref{fig:EO1} and \ref{fig:EO2}, 
the vertices of the green points represent the contact points' position. 
In both figure, (a) represents the case where there was one region of contact points, 
and we call these data sets as ``RPP1" and ``Liver1". 
And the contact points' positions in the case where there were six contact point regions are shown in (b), 
and we call these data sets with these contact points' setting as ``RPP6" and ``Liver6".

To obtain the data set of RPP1, 
deformation was performed by forcibly displacing the contact point in a rectangular parallelepiped 
of 204.8 mm $\times$ 204.8 mm $\times$ 102.4 mm centered on the initial position of the contact point. 
With a closest sample point interval of 5.12 mm, 
we transformed the model into a total of 35,301 patterns. 
The maximum displacement of the contact point from the initial position was 153.6 mm. 
The deformation data were acquired as the displacement from the initial position 
of the vertex coordinates with three-dimensional real values for each vertex.

For the data sets of RPP6, 
the sampling range was determined based on the positional relationship 
between the fixed point and the contact point and the normal vector of the object surface 
at the contact point.  
The coordinates of the sampling points ($x,y,z$) satisfied the following conditions, 
assuming that the distance from the center of the fixed points to the center of the contact points 
is $l$, the direction vector is $\mathbf{v}_\mathrm{fc}$, 
and the average vector of the normal vectors of all contact points is $\mathbf{v}_\mathrm{nv}$.

\begin{itemize}
 \item 	Inside the spheroid around $\mathbf{v}_\mathrm{fc}$ surrounding the contact point
 \item 	The radius of the ellipsoid is $0.05l$ for the $\mathbf{v}_\mathrm{fc}$ direction, $0.2l$ for the direction orthogonal to $\mathbf{v}_\mathrm{fc}$
 \item 	A vector from the contact point to ($x, y, z$) forms an acute angle with $\mathbf{v}_\mathrm{nv}$
\end{itemize}

These settings were used to obtain data because the elastic object is deformed substantially 
by rotation around a fixed point. 
A greater nonlinear deformation occurs when the elastic object is rotated 
around a fixed point compared with expansion and contraction. 
Considering the deformation of a real object, 
the direction of applied force was determined based on the normal vector direction of the surface 
of the elastic object, 
because many actions pulling in the direction perpendicular to the surface are considered, 
rather than pushing in the direction toward the fixed point.

\begin{figure}[!t]
\centering
\includegraphics[width=80mm]{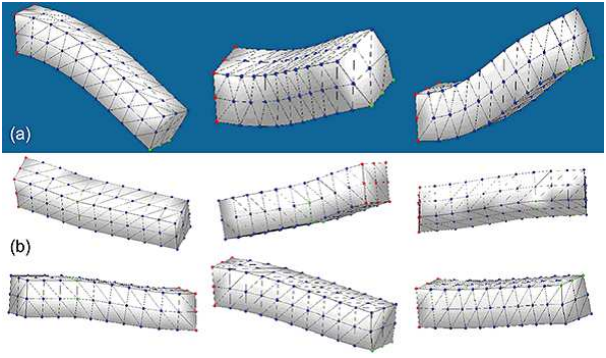}
\caption{Example of shape after deformation for rectangular parallelepiped model. 
(a) and (b) represent samples of the data sets of RPP1 and RPP6, respectively. }
\label{fig:ObtainedData1}
\end{figure}

For each region of contact points, 
sample points were equally arranged with a nearest neighbor sample point interval of $0.01l$. 
The number of sample points was somewhat different for each region of contact points, 
with an average of approximately 4,200. 
The sample point interval was defined in proportion to $l$ because the distance 
from the fixed point varies depending on the position of the contact point's area, 
and there was a need to move the contact points to a small area in which the contact points were
located close to the fixed point. 
This was necessary because applying a displacement so large 
that an actual elastic body would tear could cause calculation errors in the finite element method. 
Some examples of acquired samples are shown in Fig. \ref{fig:ObtainedData1}. 

Since the mesh structure of the liver model is more complicated 
than that of a rectangular parallelepiped model, impossible deformation was caused by
setting the sampling range inside the rectangular parallelepiped, as 
in the case of a rectangular parallelepiped model, when there was one area of contact points. 
Therefore, sample points were placed inside the ellipsoid to obtain the data set Liver1. 
Unlike the rectangular parallelepiped model, 
the fixed points for the liver model were located near the center of the whole shape, 
not at the end of the object, 
so the reference length $l$ for the size of the ellipsoid was regarded 
as the longest diameter of the liver regardless of the position of the contact point area.

\begin{figure}[!t]
\centering
\includegraphics[width=80mm]{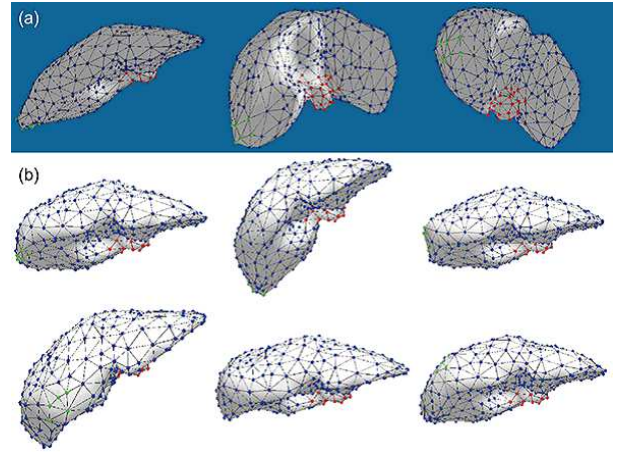}
\caption{Example of shape after deformation for liver model. 
(a) Liver1. (b) Liver6. }
\label{fig:ObtainedData2}
\end{figure}

For the data set of Liver1, 
the radius of the spheroid around $\mathbf{v}_\mathrm{fc}$ was set to $0.2l$ 
for the $\mathbf{v}_\mathrm{fc}$ direction and $0.3l$ for the direction perpendicular to the $\mathbf{v}_\mathrm{fc}$. 
The nearest neighbor sampling point interval was set to $0.01l$, which corresponded to 2.21 mm in real space. 
There were 6,089 samples in total. 
The maximum displacement of the contact point from the initial position was 66.4 mm. 
When sampling of the data set of Liver6, 
the radius of the ellipsoid was set to $0.05l$ for the $\mathbf{v}_\mathrm{fc}$ direction and 
$0.2l$ for the direction perpendicular to the $\mathbf{v}_\mathrm{fc}$, 
and the nearest neighbor sample point interval was set to $0.01l$. 
To select a sample point based on a normal vector, 
an average vector of normal vectors of the contact points was used in the case where the contact point region 
was five among six areas. 
In the other case when the contact points were located on the back of the liver, 
the normal vector in the direction toward the front side was used 
for simulating the action of turning over the liver during surgery. 
The number of samples was somewhat different for each area of contact points, 
but was approximately 4,200 on average.
Fig. \ref{fig:ObtainedData2} shows examples of the deformed shape for the data sample. 
For these deformed shapes, displacement from the initial shape was used as experimental data.

%||||||||||||||||||||||||||||||||||||||||||||||||||||||||||||||||||||||||||||||||||||||
\section{Experiments and Evaluation}
\label{sec:Result}
First, the number of nodes in the middle layers of the neural network was examined. 
In addition, we observed the test error transition during the training 
and local positional error of the shape after deformation. 
Next, we examined the effectiveness of the method for the deformed data with multiple regions of contact points. 

We carried out 5-fold cross-validation to evaluate the performance of our method. 
Namely, all of the data in one data set were randomly divided into five data sets of approximately equal size, 
one of which was used as test data, while the remaining data were used as training data. 
The data set used as test data was changed five times, 
and the estimation error across these five trials was averaged to estimate the error of one session. 

Evaluation of the error was performed by transforming the difference 
between the output and the target using the test data to the actual size, in mm shown in Fig. \ref{fig:SchematicDiagram}. 
This difference corresponds to the difference between the coordinates of all the $N$ vertices. 
The overall error was evaluated using the root mean square error (RMSE)  
at all the $N$ vertices calculated using all test data. 
For the local positional error, 
the difference in coordinates was obtained for each vertex for each test sample. 

The regularization coefficient, which was set for the weight to be optimized, was 
$\lambda_1 = \lambda_2 = \lambda_3 = 0.1$ in equation (\ref{eq:J}), 
and the initial value of the weight was randomly determined in every trial 
based on the matrix size of the weight. 
Implementation of the neural network was programmed using MATLAB (R2014b, MathWorks) 
on a personal computer.

%----------------------------------------
\subsection{Examination on the number of nodes for hidden layers}
\label{ssec:hidden}
First we set the number of observation points to three for the rectangular parallelepiped model. 
We used the edges and corners as observation points, 
because we assumed they would be relatively easy feature points to detect from images of a real elastic body. 
Since there were three observation points, 
the number of nodes $M_1$ in the input layer of the neural network in Fig. \ref{fig:NN} was $3 \times 3 = 9$, 
and the number of nodes $M_4$ in the output layer was $(99-9) \times 3 = 270$ 
to exclude only the fixed point. 

The number of nodes in the two hidden layers varied from two to 150 at two intervals, 
and training was performed in each case using the data set of RPP1. 
The training was repeated in five epochs with 10 iterations for each mini batch. 
There were 28 mini batches, with 1,000 data samples in each.
There were 1,400 iterations in total. 
During one trial, the neural network was trained using 28,241 data samples, which was 80\% of all data. 
The estimation accuracy was tested using 7,060 data samples, which was 20\% of all data, 
and this was repeated five times. 

For the liver model, we set nine vertices as observable points, and $M_1$ was $9 \times 3 = 21$. 
The number of nodes for the output layer $M_4$ was $(329-34) \times 3 = 885$. 
This number of observation points was determined based on the number of vertices for estimation 
corresponding to the percentage of observation points in the rectangular parallelepiped model. 
We examined cases in which the number of nodes in the two hidden layers varied 
from five to 400 at five intervals using the data set of Liver1. 
There were 20 training epochs, with 10 iterations for nine mini batches of 500 data samples each. 
Thus, there were 1,800 iterations in total. 
We employed a smaller mini batch size for the liver model compared with the rectangular parallelepiped model 
because the data size of one data sample was larger than that for the rectangular parallelepiped model.

\begin{figure}[!t]
\begin{minipage}[t]{\columnwidth}
\centering
\includegraphics[width=70mm]{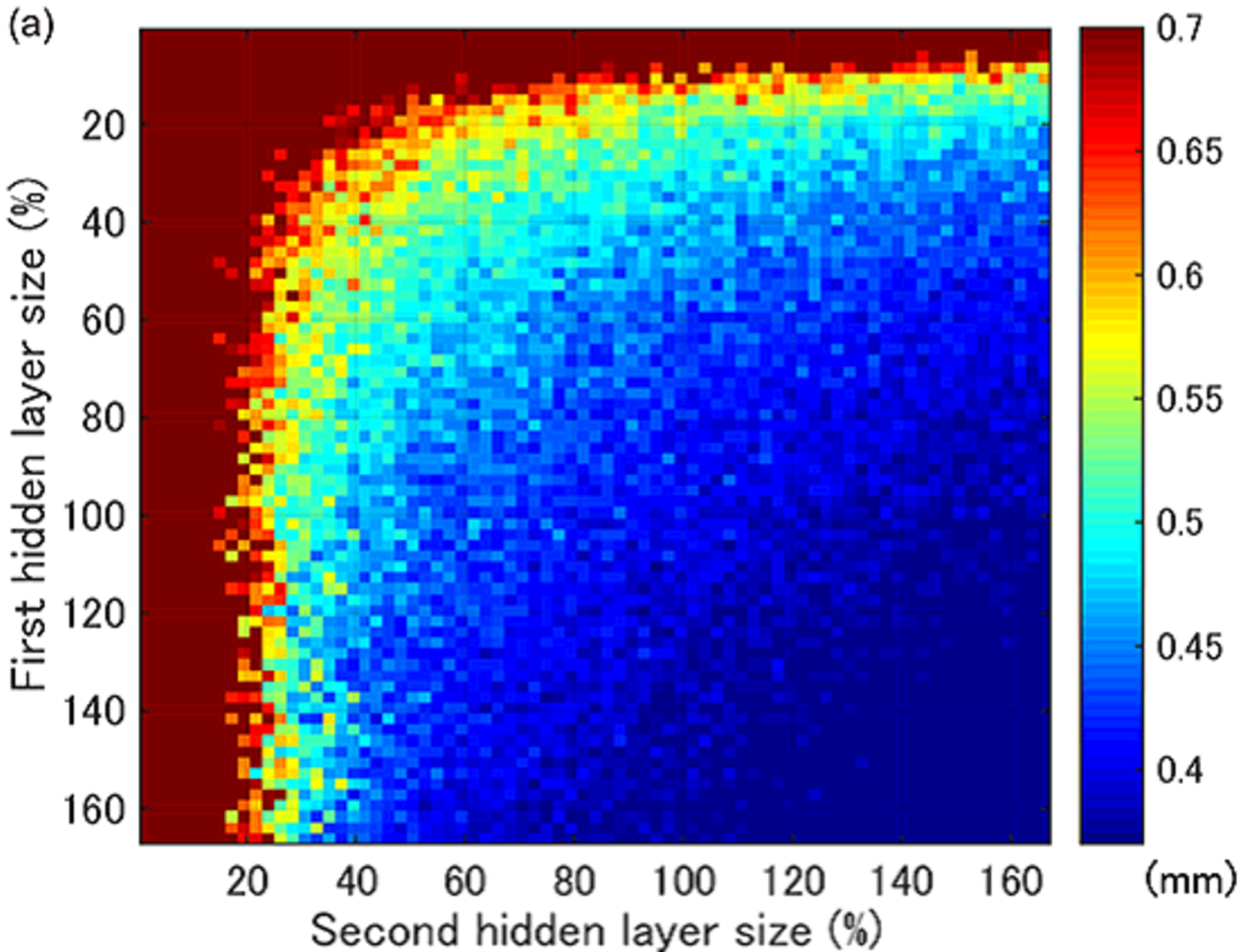}
\end{minipage}\\%
\begin{minipage}[t]{\columnwidth}
\centering
\includegraphics[width=70mm]{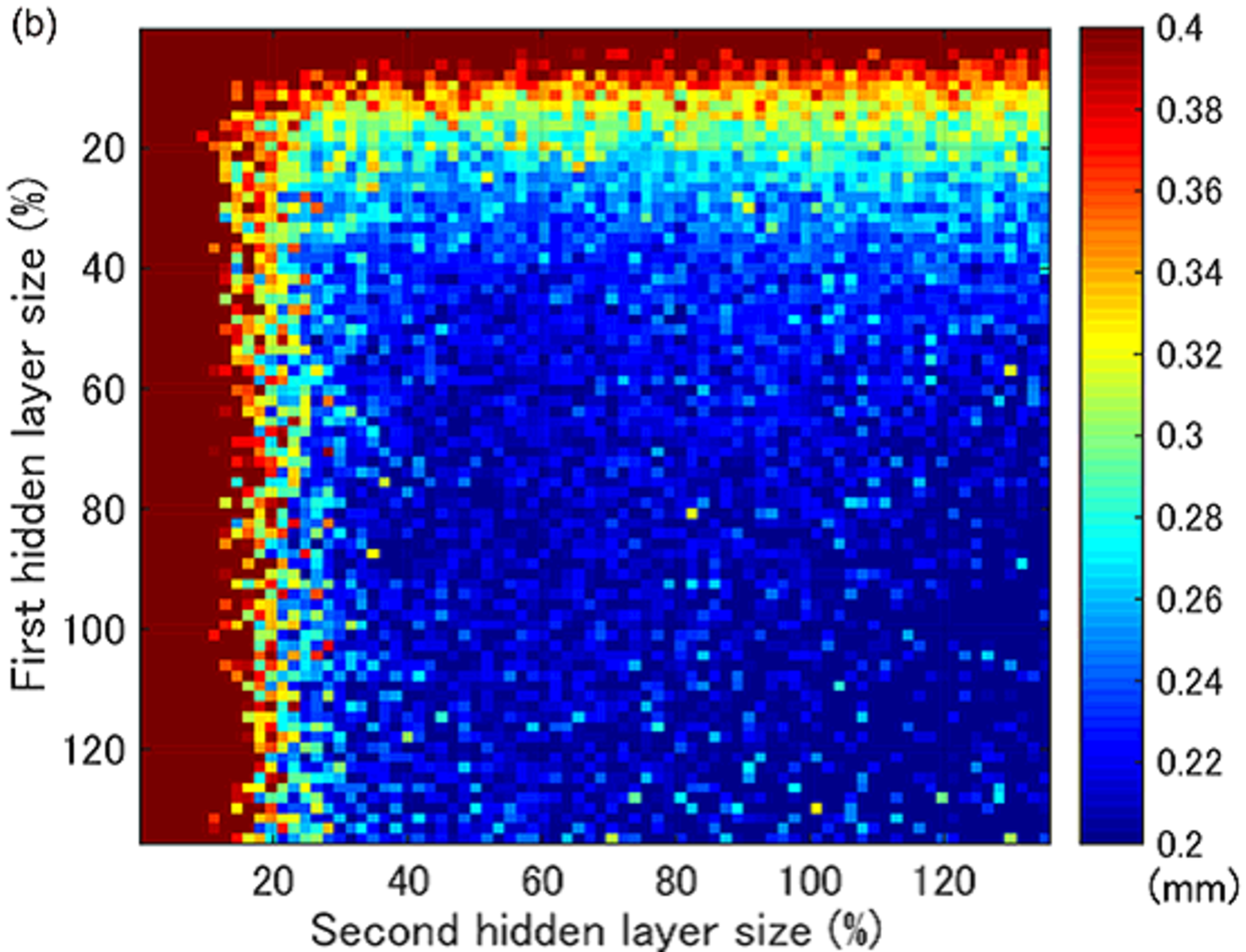}
\end{minipage}
\caption{Relationship between the number of hidden layer nodes and estimation accuracy. 
The number of hidden layer nodes is represented by the percentage of hidden layers size to the number of vertices $N$. 
Greater error is shown in red, and less error is shown in blue. 
(a) shows the results of the rectangular parallelepiped model 
and (b) shows the liver model results. }
\label{fig:HLS}
\end{figure}

Fig. \ref{fig:HLS} shows the RMSE estimated by the test data, 
with the color corresponding to the volume of error. 
A case with low error and high estimation accuracy is shown in blue, 
and a case with high error and low estimation accuracy is shown in red. 
Two axes represent the percentage of hidden layers size to the number of vertices $N$. 
In the verification range in this experiment, 
the error tended to decrease as the number of nodes increased. 
Therefore, deformation can be estimated more accurately 
by increasing the number of nodes in the hidden layers. 
However, when the number of nodes is large, the number of elements of the weights to be optimized is also large. 
As the number of elements of the weight increased, 
the time required for training and estimation increased. 
Therefore, considering real-world applications of the method, 
it may be appropriate to set a relatively small number of nodes within the range 
at which the error is as small as possible.
Fig. \ref{fig:HLS} shows that,
the RMSE was small enough when the two hidden layers were comparable in size to the number of vertices $N$. 
Therefore, in the subsequent experiments, 
the number of nodes in the two hidden layers was set as the number of vertices $N$.

%----------------------------------------
\subsection{Transition of RMSE and local positional error}
\label{ssec:RMSE_LPE}
We investigated changes in the estimation accuracy with respect to the number of training iterations 
using the data sets of RPP1 and Liver1. 
In the rectangular parallelepiped model, 
there were 90 nodes in each of the two hidden layers, 
and three observation points. 
For each of the 28 mini batches of 1,000 data samples, there were 10 iterations. 
We repeated 100 epochs, and the total number of iterations was 28,000. 
In the liver model, 
there were 295 nodes for each hidden layer, 
and there were nine observation points. 
For nine mini batches of 500 data samples, 
we performed 40 iterations each. 
The total number of iterations with 100 epoch repeats was 36,000. 
Each of the above training and test set was conducted 5 trials with different test samples for one session. 
And we repeated the session 10 times. 

\begin{figure}[!t]
\begin{minipage}[t]{\columnwidth}
\centering
\includegraphics[width=80mm]{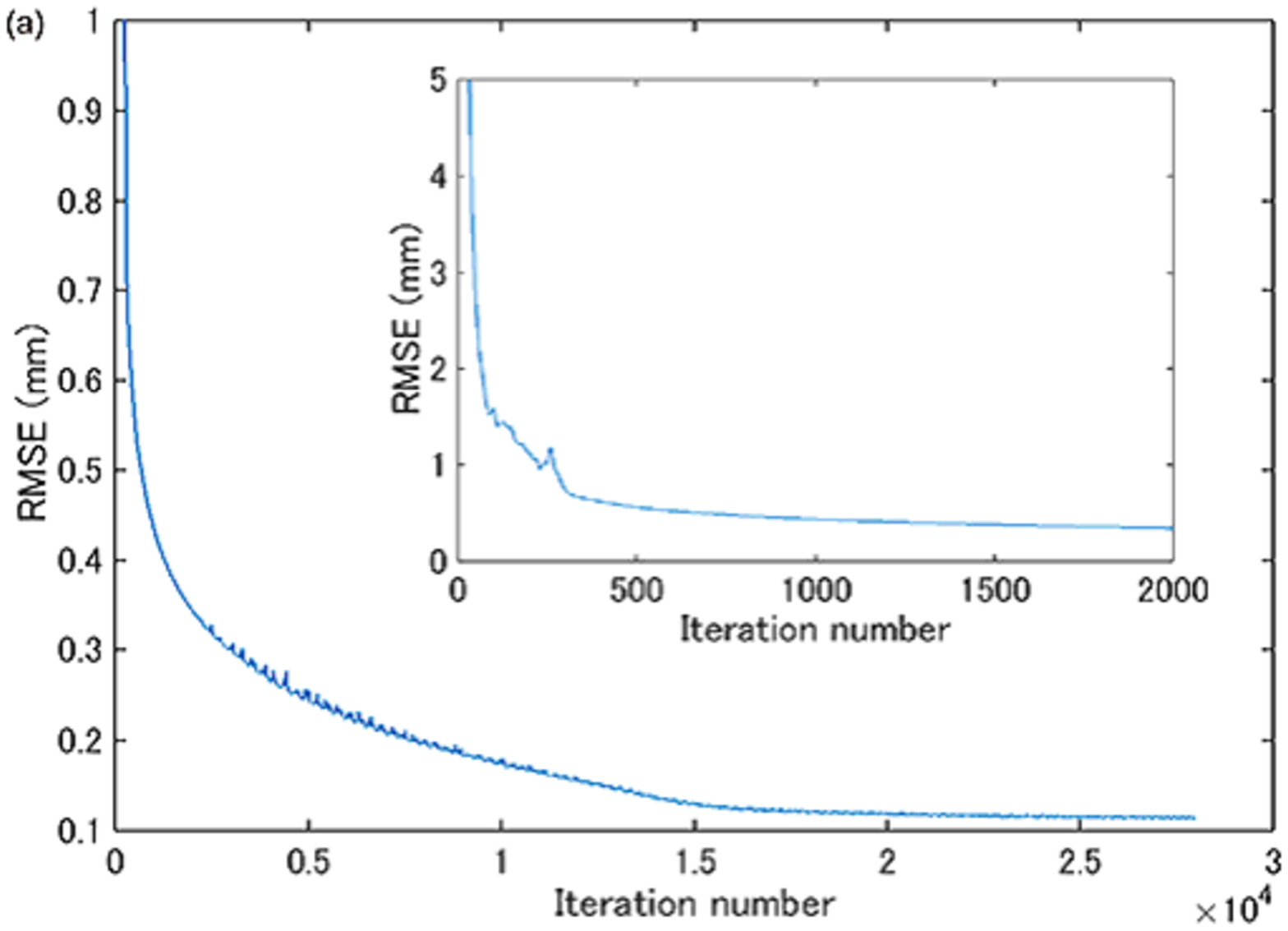}
\end{minipage}\\%
\begin{minipage}[t]{\columnwidth}
\centering
\includegraphics[width=80mm]{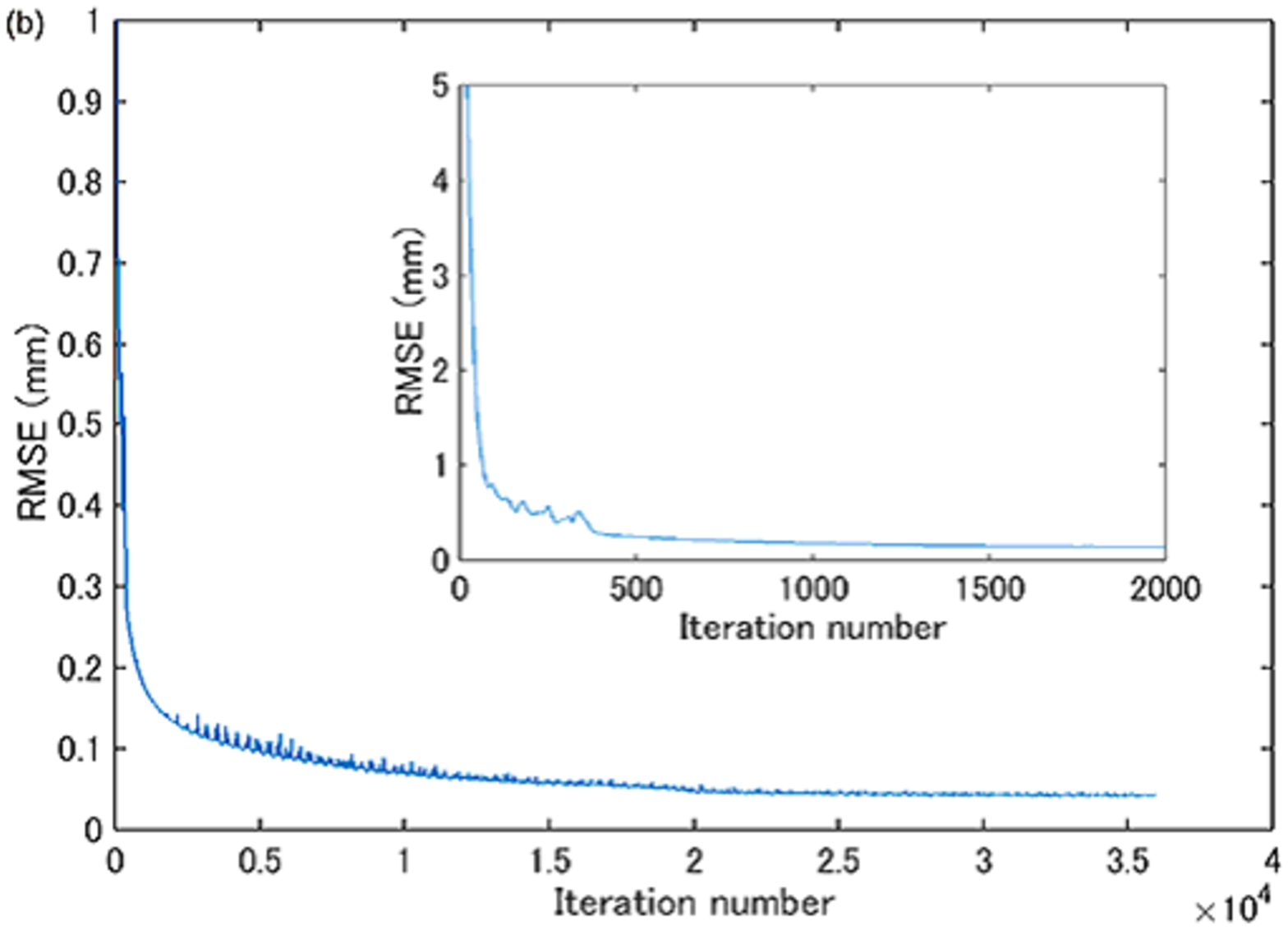}
\end{minipage}
\caption{The transition of RMSE over the number of training iterations. 
(a) refers to the rectangular parallelepiped model, and (b) refers to the liver model.}
\label{fig:RMSE_Itr}
\end{figure}

Fig. \ref{fig:RMSE_Itr} shows the transition of RMSE calculated using test data with respect to the number of iterations. 
The results consisted of the averaged transition of the RMSE across 10 sessions. 
The graph showing up to 2,000 iterations is enlarged in the figure. 
Since it was optimized using the stochastic gradient method, 
the RMSE did not decrease monotonically, but partially increased. 
Overall, as the number of training iterations increased, 
the RMSE decreased. 
The enlarged view of a small number of training iterations 
showed that the RMSE roughly converged with approximately 1,000 iterations, 
which was less than the number of iterations employed in the previous section.

\begin{figure}[!t]
\begin{minipage}[t]{\columnwidth}
\centering
\includegraphics[width=80mm]{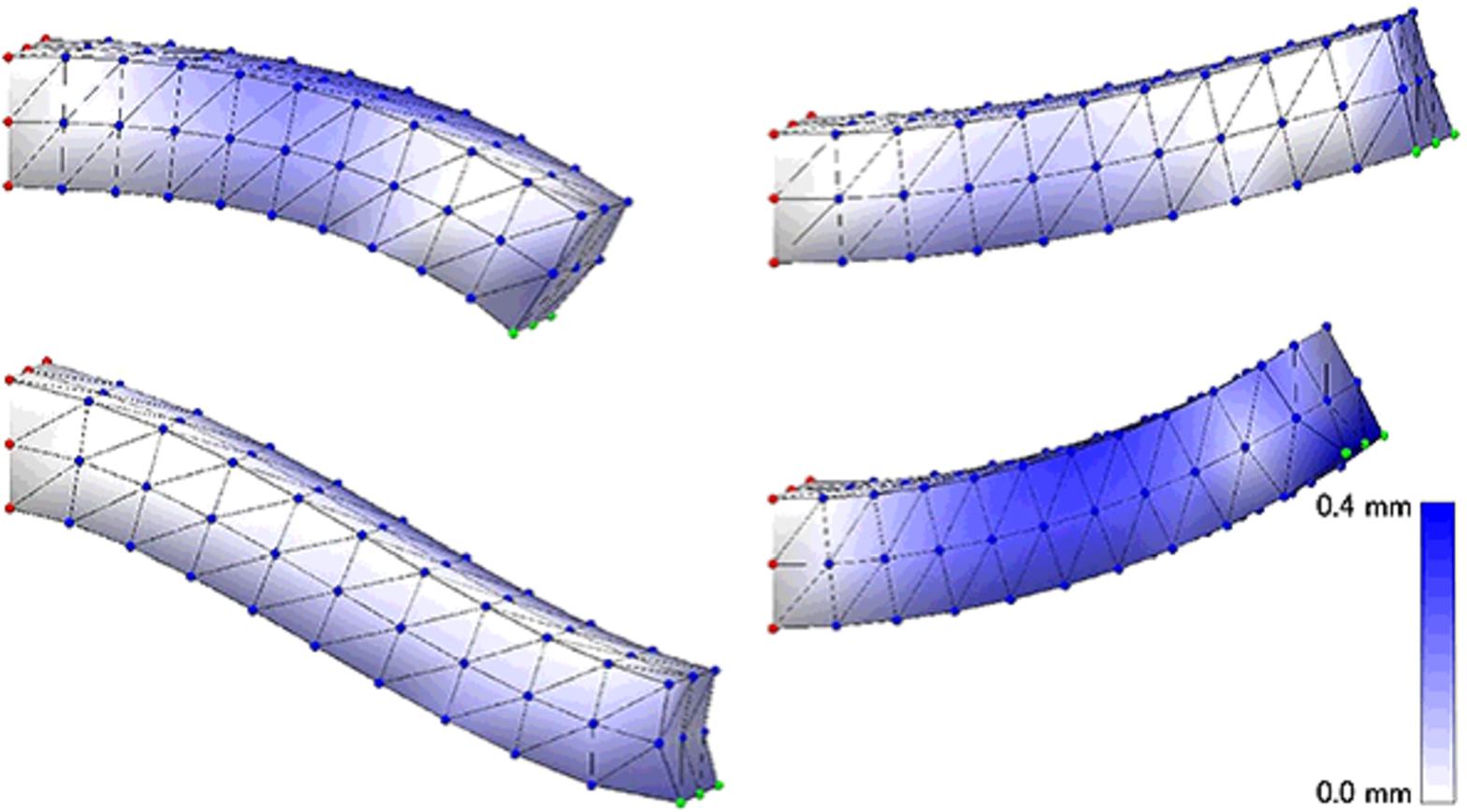}
\end{minipage}\\%
\begin{minipage}[t]{\columnwidth}
\centering
\includegraphics[width=80mm]{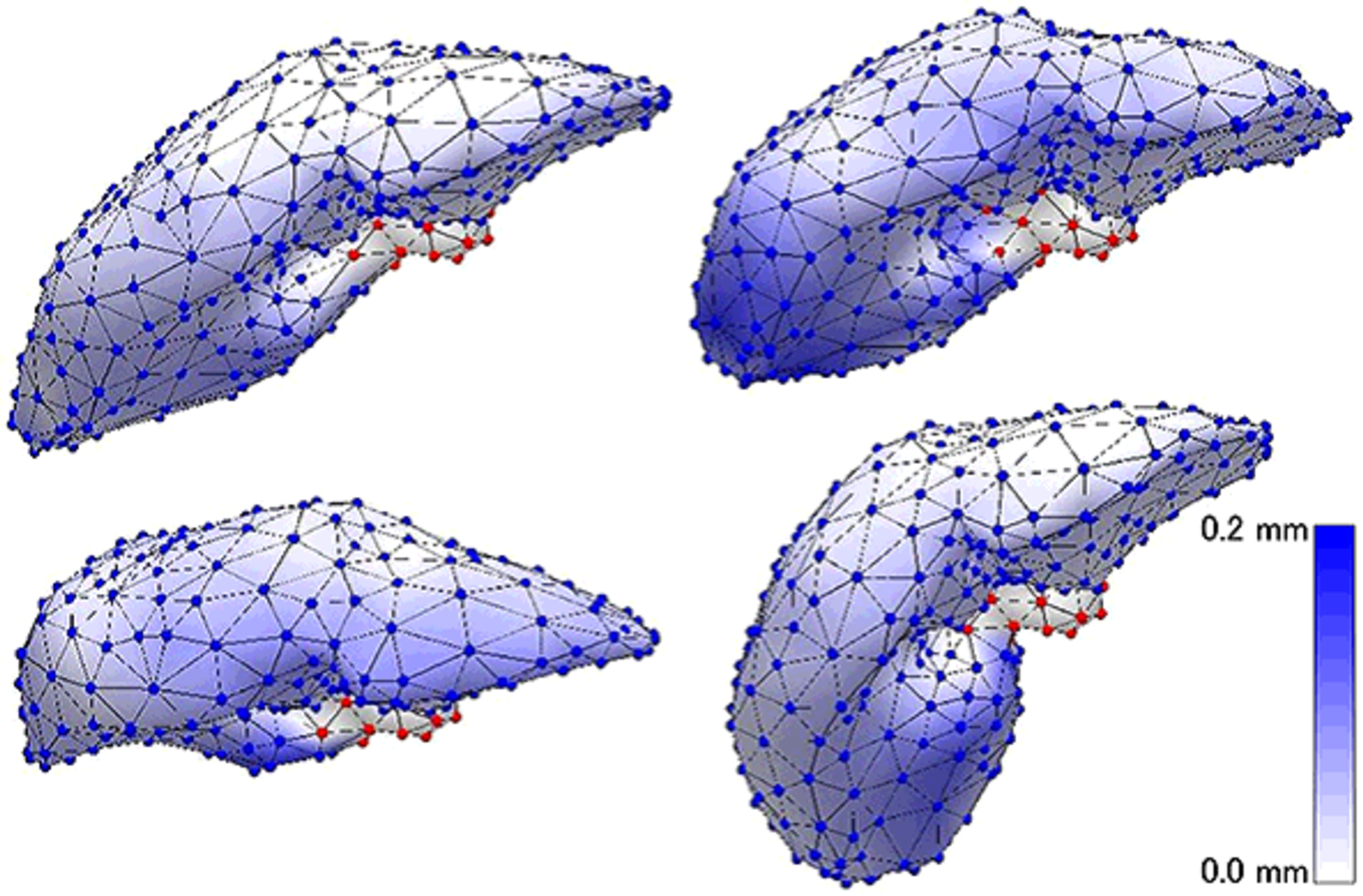}
\end{minipage}
\caption{Local positional error of shape estimation for test data. 
Darker color represents larger error.}
\label{fig:Shape}
\end{figure}

Some examples of the estimated shape using the weight of the neural network 
after the final training are shown in Fig. \ref{fig:Shape}. 
The local positional error of each vertex is displayed as a color map on the surface. 
The surface was colored dark blue around the vertices, 
where the difference of the vertex coordinates was large. 
For the rectangular parallelepiped model, 
the error was relatively large around the position where the shape curved substantially.
However, the results showed that the displacement could be estimated with less error as a whole, 
even in the region where the object deformed nonlinearly and part of it was stretched.

The average of 10 sessions of the final RMSE was 0.114 mm for the rectangular parallelepiped model 
and 0.041 mm for the liver model. 
That is, the percentage of error against the maximum value of the contact point displacement 
for obtaining the deformed shape was 0.074\% for the rectangular parallelepiped model 
and 0.062\% for the liver model. 
The results revealed that the displacement of all vertices excluding the fixed points 
could be estimated with the small error on average from the displacement of observation points 
which was only 3\% of all vertices to be estimated.

%----------------------------------------
\subsection{Influence of multiple contact region}
\label{ssec:multiple}
Using the data sets of RPP6 and Liver6, 
the estimation performance was examined when the region of contact points had multiple area. 
All six sets of data with different regions of contact points were mixed. 
We used 80\% of all data for training, and testing with 20\% of data was repeated five trials. 
And this session repeated 10 times. 
For 20 mini batches of 1000 data samples and 40 mini batches of 500 data samples, 
we performed 15 and ten iterations each using RPP6 and Liver6, respectively. 
The total number of iterations with 100 epoch repeats was 30,000 and 40,000. 
We conducted 5-fold cross-validation including 5 trials with different test samples for one session 
and we repeated the session 10 times. 

The results were compared with those of the experiment in the section \ref{ssec:RMSE_LPE}. 
Table \ref{tab:RMSE_LPE} shows the estimation results and the data set information. 
In the case of ``RPP1" and ``Liver1", the results are shown in section \ref{ssec:RMSE_LPE}. 
``Mean RMSE" was obtained by averaging the squared error of all vertices in 10 sessions 
estimated using all test data in each trial. 
The percentage is the ratio of ``Mean RMSE" to the ``Max displacement". 
``Mean of max LPE" indicates the average value of maximum of the local positional error for each vertex of each test data sample. 
Each maximum value of the local positional error was obtained from one sample of test data. 
The percentage represents the ratio of ``Mean of max LPE" to the ``Max displacement". 
``Max displacement" indicates the maximum displacement of contact points in the data set 
used for training and testing. 
``Observation point" refers to the percentage of the number of observation points to the number of vertices $N$. 
The number of observation points is also shown. 
``$N$ vertices" indicates the number of vertices needed to estimate the displacement. 
The vertex to be estimated for the displacement corresponds to the vertex excluding the fixed point from all vertices. 
``Number of all data sample" refers to the total number of data samples used for the training and test. 

\begin{table*}[!t]
\renewcommand{\arraystretch}{1.3}
\caption{Estimation error using each data set and the data set information.}
\label{tab:RMSE_LPE}
\centering
\begin{tabular}{|c||c|c|c|c|c|c|}
\hline
 & RPP1 & RPP6 & Liver1 & Liver6 \\
\hline
\hline
Mean RMSE (mm (\%)) & 0.114 (0.074) & 0.283 (0.550) & 0.041 (0.062) & 0.275 (0.621) \\
\hline
Mean of max LPE (mm (\%)) & 0.332 (0.216) & 1.288 (2.502) & 0.166 (0.250) & 1.698 (3.836) \\
\hline
Max displacement (mm) & 153.6 & 51.5 & 66.4 & 44.3 \\
\hline
Observation point (\% (number)) & 3.33 (3) & 3.33 (3) & 3.05 (9) & 3.05 (9) \\
\hline
$N$ vertices (number) & 90 & 90 & 295 & 295 \\
\hline
Number of all data sample & 35301 & 25146 & 6089 & 25140 \\
\hline
\end{tabular}
\end{table*}

In the case where there was one area of contact points, 
comparison of actual size of ``Mean RMSE" revealed that 
the liver model produced less error than the rectangular parallelepiped model. 
For data with one region of contact points, 
the maximum displacement of contact points was around three times larger in the RPP1 compared with the data set of Liver1. 
The percentage of ``Mean RMSE" revealed no large difference between the two models. 
For data with multiple regions of contact points, 
the maximum displacement of contact points and the number of data samples were similar in both models. 
The error in the data set of RPP6 and Liver6 were comparable. 
The variation of deformation patterns increases in the data set with multiple contact region. 
The average error was larger than the case when the contact points were determined in the single region. 

\begin{figure}[!t]
\centering
\includegraphics[width=80mm]{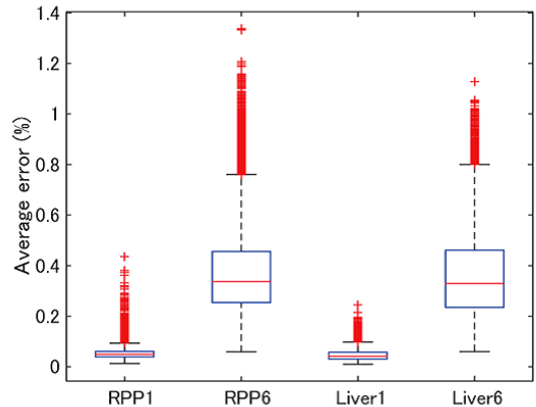}
\caption{Average value of local positional error for each test data sample. 
The error is represented by the percentage to the maximum displacement of the contact points for each data set. }
\label{fig:MeanLPE}
\end{figure}

The average error of the local positional error for each test data in one session was shown by a box plot in Fig. \ref{fig:MeanLPE}. 
The error was represented by the percentage to the maximum displacement of the contact points for each data set. 
The median ratio of the average error was slightly smaller in the liver model than that of the rectangular parallelepiped model 
in both cases of single and multiple contact regions. 
The results imply that the displacement of all vertex would be estimated by our method in the average error of 0.33 \% 
when the data set has contact points in multiple regions. 

\begin{figure}[!t]
\centering
\includegraphics[width=80mm]{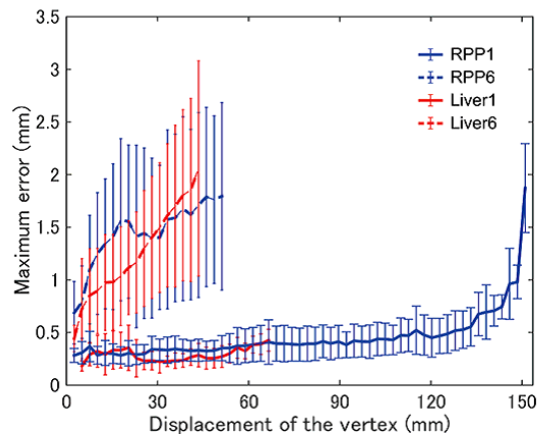}
\caption{The relationship between the maximum value of local positional error for each test data sample 
and the displacement of that vertex.}
\label{fig:MaxLPE}
\end{figure}

In addition, the relationship between the maximum value of the local positional error for each test data in one trial 
and the displacement of the vertex having the maximum error in the test data 
is shown in Fig. \ref{fig:MaxLPE}. 
The mean value of the maximum error for each displacement of the vertex is displayed by line with error bar of the standard deviation. 
The blue and red lines demonstrate the results of the rectangular parallelepiped model and the liver model, respectively. 
The solid line is for the case of the single contact region 
and the dashed line is for the case of the multiple contact regions. 
The maximum error increased following the size of displacement in the case of the multiple contact regions. 
When the contact region was single area, 
the maximum error more correlated with the amount of the displacement in the larger range of the displacement. 

%||||||||||||||||||||||||||||||||||||||||||||||||||||||||||||||||||||||||||||||||||||||
\section{Discussions}
\label{sec:Discussion}
When we examined the hidden layer size for the liver model, 
the RMSE had relatively large value in some cases in the area 
where the size of two hidden layers was larger than 100 \%. 
It is considered that the convergence was too slow in some cases, 
because the number of parameters of weight to be optimized was too large. 
Although the expressiveness of the neural network and the ability to match the data increases 
when the number of elements of the weights to be optimized is large, 
too slow convergence obstructs real-world applications of the method. 

In this study we employed three and nine observation points when we used the rectangular parallelepiped model and the liver model, respectively. 
These number of observation points are corresponding to around 3\% of all vertices to be estimated the displacement, and 
we determined these number by trial and error. 
For practical applications, 
it is more feasible to estimate the overall deformation, 
even if the number of observation points is small. 
When an actual organ is deformed and the deformation is measured with a camera,
regions with a characteristic shape are limited. 
We assume that deformation can be tracked, and displacement can be measured only in such characteristic regions. 
Furthermore, the event considered to be the same case as above paragraph was observed when we examined the number of observation points. 
Since the number of nodes in the input layer was three times the number of observation points, 
increasing the number of observation points increased the number of nodes in the input layer of the neural network. 
When the number of observation points was too large, 
the number of elements of the optimization parameter increased too much and the convergence became too slow. 
%In addition, since the number of nodes in the middle layer was fixed, 
%even though the number of nodes in the input layer increased, 
%the expressive power of the neural network appeared to be limited. 
%When the number of observation points was relatively small, 
%as the number of observation points increased, 
%input information about the deformation of the elastic body 
%and the expressive power of the neural network increased, and the RMSE decreased. 

The value of 75th percentile of the liver model was larger 
than that of the rectangular parallelepiped model as shown in Fig. \ref{fig:MeanLPE}. 
Since the mesh structure around the fixed point was more complicated in the liver model 
than the rectangular parallelepiped model, 
it appears that the area with large variation caused a large error in some samples. 
The median value of the error of 0.33 \% corresponds to 0.146 mm in the liver model. 
Fig. \ref{fig:MaxLPE} shows the maximum error using the data set of Liver6 was largest among examinations with four data sets. 
This data set could be considered having a lot of variation due to the complexity of the mesh structure. 
Even under this difficult estimation condition, 
it was possible to estimate within the maximum error of 2.06 mm on average. 
This result shows the maximum error, 
guaranteeing that a partially large error falls within this error range. 
The interval between the mesh vertices of the liver model was approximately 15 mm. 
%These findings suggest that the maximum error appeared to be within an acceptable range.
By comparison with this interval, the maximum error would not be so significantly big.

As a future work, we consider the transfer learning of the neural network, which was trained with a standard model of the targeted internal organ. 
For example, we make the location of vertices of the mesh model matching to the original model 
and apply the trained neural network through this research to the other liver model constructed by CT data of the other person. 
Furthermore for practical applications, we consider to apply this method to the actual internal organ. 
Since whole deformation could not be observed in the actual internal organ, 
it seems to need to train the neural network from partial observation information, which has various location in each subject. 
It is considered to combine both training using the mesh model which can represent whole deformation 
and using the actual organ which can provide only partial information of deformation.

%||||||||||||||||||||||||||||||||||||||||||||||||||||||||||||||||||||||||||||||||||||||
\section{Conclusion}
\label{Fin}
We proposed a method for estimating deformation of a whole elastic object 
based on partial observation, using a neural network and showed the feasibility of estimation. 
A rectangular parallelepiped model and a human liver model reconstructed from CT data were used as elastic objects to evaluate the method. 
The data samples were obtained by the nonlinear finite element method 
in the condition that the initial shape of the elastic object was known. 
For training the neural network, 
displacement of a very small number of observation points was taken as input, 
and displacement of all vertices, including non-observation points and excluding fixed points, was the output. 

In the current experiment, we investigated the estimation performance of our proposed method, 
examining the hidden layer size, the number of iterations of optimization, 
and the position of contact points. 
We selected the vertices considered to be most suitable for tracking displacement, 
and set the number of nodes in two hidden layers in accord with the number of mesh vertices arranged in the elastic object. 
We showed that the estimation accuracy of the position error of vertex coordinates was 0.041 mm and 0.275 mm on average 
from only around 3 \% observations among all vertices using the liver model with single and multiple contact regions, respectively. 
The current results demonstrated that the proposed method enables estimation of deformation with an error of much less than 1 mm, 
which is tolerable in the context of supporting surgical operations. 
The proposed method has potential applications for deformation estimation of internal organs.

%||||||||||||||||||||||||||||||||||||||||||||||||||||||||||||||||||||||||||||||||||||||
\section*{Acknowledgment}
This research was supported by 
the Center of Innovation (COI) Program from the Japan Science and Technology Agency (JST) 
and Japan Society for the Promotion of Science (JSPS) Grant-in-Aid for Young Scientists (B) Grant Number JP16K16407.

%||||||||||||||||||||||||||||||||||||||||||||||||||||||||||||||||||||||||||||||||||||||
\section*{Conflict of interest statement}
None Declared.

\ifCLASSOPTIONcaptionsoff
  \newpage
\fi

\bibliographystyle{IEEEtran.bst}
\bibliography{Bibliography}

\begin{thebibliography}{10}
\providecommand{\url}[1]{#1}
\csname url@samestyle\endcsname
\providecommand{\newblock}{\relax}
\providecommand{\bibinfo}[2]{#2}
\providecommand{\BIBentrySTDinterwordspacing}{\spaceskip=0pt\relax}
\providecommand{\BIBentryALTinterwordstretchfactor}{4}
\providecommand{\BIBentryALTinterwordspacing}{\spaceskip=\fontdimen2\font plus
\BIBentryALTinterwordstretchfactor\fontdimen3\font minus
  \fontdimen4\font\relax}
\providecommand{\BIBforeignlanguage}[2]{{%
\expandafter\ifx\csname l@#1\endcsname\relax
\typeout{** WARNING: IEEEtran.bst: No hyphenation pattern has been}%
\typeout{** loaded for the language `#1'. Using the pattern for}%
\typeout{** the default language instead.}%
\else
\language=\csname l@#1\endcsname
\fi
#2}}
\providecommand{\BIBdecl}{\relax}
\BIBdecl

\bibitem{Shu2013}
Z.-B. Shu \emph{et~al.}, ``Laparoscopic versus open resection of gastric
  gastrointestinal stromal tumors,'' \emph{Chinese J. Cancer Res.}, vol.~25,
  no.~2, pp. 175--182, 4 2013.

\bibitem{Fuentes2014}
M.~N. Fuentes \emph{et~al.}, ``Complications of laparoscopic gynecologic
  surgery,'' \emph{J. Soc. Laparoendosc. Surgeons}, vol.~18, no.~3, p.
  e2014.00058, 7 2014.

\bibitem{Meadows2002}
M.~Meadows, ``Robots lend a helping hand to surgeons,'' \emph{{U}.{S}. Food and
  Drug Admin. Consum. Mag.}, vol.~36, no.~3, 6 2002.

\bibitem{Malti2014}
A.~Malti and A.~Bartoli, ``Combining conformal deformation and cook-torrance
  shading for 3-{D} reconstruction in laparoscopy,'' \emph{{IEEE} Trans.
  Biomed. Eng}, vol.~61, no.~6, pp. 1684--1692, 6 2014.

\bibitem{Qin2013}
Y.~Qin, H.~Hua, and M.~Nguyen, ``Multiresolution foveated laparoscope with high
  resolvability,'' \emph{Opt. Lett.}, vol.~38, no.~13, pp. 2191--2193, 7 2013.

\bibitem{Puerto2014}
G.~A. Puerto-Souza and G.-L. Mariottini, ``Wide-baseline dense feature matching
  for endoscopic images,'' \emph{Proc. {PSIVT} 2013: Image and Video Technol.},
  pp. 48--59, 10 2014.

\bibitem{Lin2015}
B.~Lin \emph{et~al.}, ``Efficient vessel feature detection for endoscopic image
  analysis,'' \emph{{IEEE} Trans. Biomed. Eng}, vol.~62, no.~4, pp. 1141--1150,
  4 2015.

\bibitem{Hagan2014}
M.~T. Hagan \emph{et~al.}, \emph{Neural network design}, 2nd~ed.\hskip 1em plus
  0.5em minus 0.4em\relax Martin Hagan, 9 2014.

\bibitem{Kecman2001}
V.~Kecman, \emph{Learning and soft computing: support vector machines, neural
  networks, and fuzzy logic models}.\hskip 1em plus 0.5em minus 0.4em\relax
  Cambridge, MA, USA: MIT Press, 2001.

\bibitem{LeCun1989}
Y.~LeCun \emph{et~al.}, ``Backpropagation applied to handwritten zip code
  recognition,'' \emph{Neural Comput.}, vol.~1, no.~4, pp. 541--551, 1989.

\bibitem{Lecun1998}
Y.~Lecun \emph{et~al.}, ``Gradient-based learning applied to document
  recognition,'' \emph{Proc. {IEEE}}, vol.~86, no.~11, pp. 2278--2324, 11 1998.

\bibitem{Belytschko2014}
T.~Belytschko \emph{et~al.}, \emph{Nonlinear finite elements for continua and
  structures}, 2nd~ed.\hskip 1em plus 0.5em minus 0.4em\relax Wiley, 1 2014.

\bibitem{FCAMRT2015}
E.~Seeram, \emph{Computed tomography: physical principles, clinical
  applications, and quality control}, 4th~ed.\hskip 1em plus 0.5em minus
  0.4em\relax Saunders, 10 2015.

\bibitem{Nakao2010}
M.~Nakao and K.~Minato, ``Physics-based interactive volume manipulation for
  sharing surgical process,'' \emph{{IEEE} Trans. Inform. Technol. Biomed.},
  vol.~14, no.~3, pp. 809--816, 5 2010.

\bibitem{Berkley2004}
J.~Berkley \emph{et~al.}, ``Real-time finite element modeling for surgery
  simulation: an application to virtual suturing,'' \emph{{IEEE} Trans. Vis.
  Comput. Graphics}, vol.~10, no.~3, pp. 314--325, 5 2004.

\bibitem{Nakao2007}
M.~Nakao \emph{et~al.}, ``Simulating lung tumor motion for dynamic
  tumor-tracking irradiation,'' in \emph{2007 {IEEE} Nucl. Sci. Symp. Conf.
  Rec.}, vol.~6, 10 2007, pp. 4549--4551.

\bibitem{Muller2002}
M.~M\"{u}ller \emph{et~al.}, ``Stable real-time deformations,'' in \emph{Proc.
  2002 {ACM} {SIGGRAPH}/Eurographics Symp. Comput. Animat.}, ser. SCA
  '02.\hskip 1em plus 0.5em minus 0.4em\relax New York, NY, USA: ACM, 2002, pp.
  49--54.

\bibitem{Kikuuwe2009}
R.~Kikuuwe, H.~Tabuchi, and M.~Yamamoto, ``An edge-based computationally
  efficient formulation of saint venant-kirchhoff tetrahedral finite
  elements,'' \emph{{ACM} Trans. Graphics}, vol.~28, no.~1, pp. 8:1--8:13, 1
  2009.

\bibitem{Suwelack2014}
S.~Suwelack \emph{et~al.}, ``Physics-based shape matching for intraoperative
  image guidance,'' \emph{Med. Phys.}, vol.~41, no.~11, p. 111901, 11 2014.

\bibitem{Morooka2010}
K.~Morooka \emph{et~al.}, ``Real-time nonlinear {FEM}-based simulator for
  deforming volume model of soft organ by neural network (in japanese),''
  \emph{{IEICE} Trans. Inform. Syst.}, vol.~93, no.~3, pp. 365--376, 3 2010.

\bibitem{Morooka2012}
K.~Morooka \emph{et~al.}, ``A method for constructing real-time {FEM}-based
  simulator of stomach behavior with large-scale deformation by neural
  networks,'' in \emph{Proc. {SPIE}}, vol. 8316, 2012, pp.
  83\,160J--83\,160J--6.

\bibitem{Morooka2013}
K.~Morooka, M.~Nakamoto, and S.~Yoshinobu, ``A survey on statistical modeling
  and machine learning approaches to computer assisted medical intervention:
  Intraoperative anatomy modeling and optimization of interventional
  procedures,'' \emph{{IEICE} Trans. Inform. Syst.}, vol.~96, no.~4, pp.
  784--797, 2013.

\bibitem{MaierHein2013}
L.~Maier-Hein \emph{et~al.}, ``Optical techniques for 3{D} surface
  reconstruction in computer-assisted laparoscopic surgery,'' \emph{Med. Image
  Anal.}, vol.~17, no.~8, pp. 974--996, 2013.

\bibitem{Santos2014}
T.~R. dos Santos \emph{et~al.}, ``Pose-independent surface matching for
  intra-operative soft-tissue marker-less registration,'' \emph{Med. Image
  Anal.}, vol.~18, no.~7, pp. 1101--1114, 2014.

\bibitem{Simpson2012}
A.~L. Simpson \emph{et~al.}, \emph{Model-assisted image-guided liver surgery
  using sparse intraoperative data}.\hskip 1em plus 0.5em minus 0.4em\relax
  Berlin, Heidelberg: Springer Berlin Heidelberg, 2012, pp. 7--40.

\bibitem{Dumpuri2010}
P.~Dumpuri \emph{et~al.}, ``Model-updated image-guided liver surgery:
  preliminary results using surface characterization,'' \emph{Prog. Biophys.
  Mol. Bio.}, vol. 103, no. 2-3, pp. 197--207, 12 2010.

\bibitem{Sakata2017}
R.~Sakata, M.~Nakao, and T.~Matsuda, ``Estimation of external forces on the
  basis of local displacement observations of an elastic body,'' \emph{Adv.
  Biomed. Eng.}, vol.~6, pp. 21--27, 2017.

\bibitem{Saito2015}
A.~Saito \emph{et~al.}, ``Deformation estimation of elastic bodies using
  multiple silhouette images for endoscopic image augmentation,'' in \emph{2015
  {IEEE} Int. Symp. Mixed and Augmented Reality}, 9 2015, pp. 170--171.

\bibitem{Greminger2003}
M.~A. Greminger and B.~J. Nelson, ``Modeling elastic objects with neural
  networks for vision-based force measurement,'' in \emph{Proc. 2003
  {IEEE}/{RSJ} Int. Conf. Intell. Robots and Syst.}, vol.~2, 10 2003, pp.
  1278--1283.

\bibitem{Aviles2014}
A.~I. Aviles \emph{et~al.}, ``A recurrent neural network approach for 3{D}
  vision-based force estimation,'' in \emph{4th Int. Conf. Image Process.
  Theory, Tools and Applicat.}, 10 2014, pp. 1--6.

\bibitem{Karimirad2013}
F.~Karimirad \emph{et~al.}, ``Modelling a precision loadcell using neural
  networks for vision-based force measurement in cell micromanipulation,'' in
  \emph{2013 {IEEE}/{ASME} Int. Conf. Adv. Intell. Mech.}, 7 2013, pp.
  106--110.

\bibitem{Karimirad2014}
F.~Karimirad, S.~Chauhan, and B.~Shirinzadeh, ``Vision-based force measurement
  using neural networks for biological cell microinjection,'' \emph{J.
  Biomech.}, vol.~47, no.~5, pp. 1157--1163, 2014.

\bibitem{Cretu2008}
A.-M. Cretu, P.~Payeur, and E.~M. Petriu, ``Neural network mapping and
  clustering of elastic behavior from tactile and range imaging for virtualized
  reality applications,'' \emph{{IEEE} Trans. Instrum. Meas.}, vol.~57, no.~9,
  pp. 1918--1928, 9 2008.

\bibitem{Cretu2010}
A.-M. Cretu \emph{et~al.}, ``Estimation of deformable object properties from
  shape and force measurements for virtualized reality applications,'' in
  \emph{2010 {IEEE} Int. Symp. Haptic Audio Visual Envir. and Games}, 10 2010,
  pp. 1--6.

\bibitem{Nair2010}
V.~Nair and G.~E. Hinton, ``Rectified linear units improve restricted boltzmann
  machines,'' in \emph{Proc. 27th Int. Conf. Mach. Learn.}, 2010.

\bibitem{Kingma2014}
D.~P. Kingma and J.~Ba, ``Adam: {A} method for stochastic optimization,''
  \emph{CoRR}, vol. abs/1412.6980, 2014.

\bibitem{Si2015}
H.~Si, ``{T}et{G}en, a delaunay-based quality tetrahedral mesh generator,''
  \emph{{ACM} Trans. Math. Softw.}, vol.~41, no.~2, pp. 11:1--11:36, 2 2015.

\bibitem{Nakao2014}
M.~Nakao \emph{et~al.}, ``Direct volume manipulation for visualizing
  intraoperative liver resection process,'' \emph{Comput. Meth. Prog. Biomed.},
  vol. 113, no.~3, pp. 725--735, 2014.

\end{thebibliography}

%\begin{IEEEbiography}{Utako~Yamamoto}
%\end{IEEEbiography}
%
%\begin{IEEEbiography}{Megumi~Nakao}
%\end{IEEEbiography}
%
%\begin{IEEEbiography}{Masayuki~Ohzeki}
%\end{IEEEbiography}
%
%\begin{IEEEbiography}{Tetsuya~Matsuda}
%\end{IEEEbiography}

\end{document}